# DISCOVERING LATENT INFORMATION BY SPREADING ACTIVATION ALGORITHM FOR DOCUMENT RETRIEVAL

Vuong M. Ngo

{vuong.cs@gmail.com}

**ABSTRACT**

*Syntactic search relies on keywords contained in a query to find suitable documents. So, documents that do not contain the keywords but contain information related to the query are not retrieved. Spreading activation is an algorithm for finding latent information in a query by exploiting relations between nodes in an associative network or semantic network. However, the classical spreading activation algorithm uses all relations of a node in the network that will add unsuitable information into the query. In this paper, we propose a novel approach for semantic text search, called query-oriented-constrained spreading activation that only uses relations relating to the content of the query to find really related information. Experiments on a benchmark dataset show that, in terms of the MAP measure, our search engine is 18.9% and 43.8% respectively better than the syntactic search and the search using the classical constrained spreading activation.*

## 1. INTRODUCTION

With rapid development of the Word Wide Web and e-societies, information retrieval (IR) has many challenges in exploiting those rich and huge information resources. Whereas, the keyword based IR has many limitations in finding suitable documents for user's queries. Semantic search improves search precision and recall by understanding user's intent and the contextual meaning of terms in documents and queries. Semantic search uses a set of techniques for exploiting knowledge from an ontology ([5], [20]).

A query is usually brief when expressing ideas of the user ([34]). To better clarify the content of the query, the query expansion method is widely used in information retrieval community ([26], [25], [4]). Query expansion often increased the recall ([31], [9]) and can also improve the precision ([24]).

Query expansion is added to the query latent information which does not appears in the query but show clearly the meaning of the query. Now there are many query expansion methods, such as: (1) Relevance Feedback ([14], [12]); (2) Search engine logs ([38], [33]); (3) Co-occurrence relation ([15], [28]); (4) spreading activation ([29], [30]).

The spreading activation method search the ontology concepts related to the concepts in the query through relations in the ontology. The activated concepts will add to the query. The additional information in accordance with the idea of users often increase the efficiency of document retrieval. In contrast, the retrieval efficiency usually reduced if additional information is not appropriate.

For example, consider the queries to find documents on the following: (1) "*cities that are tourist destinations of Thailand*"; (2) "*Jewish settlements are built in the east of Jerusalem*"; and (3)

"*works of Ernest Hemingway*". In the first query, *Chiang Mai* and *Phuket* should be added to the query because they are belonging to class *City* and are tourist destinations of *Thailand*. Adding query, the other famous tourist cities such as *Jakarta* and *New York* is not appropriate, because the two cities are not in *Thailand*. In the second query, the Jewish settlements built in the east of the city of *Jerusalem* as *Beit Orot* and *Beit Yehonatan* should be added to the query. In the third query, *The Old Man and the Sea* and *A Farewell to Arms* should be added to the query because they are the works of writer *Ernest Hemingway*.

The previous spreading activation used all relations of a node in an ontology, although most of the relations are not mentioned in the user's query. So, there are many unsuitable terms added into the query. In this paper, we expand query latent entity named by spreading on the ontology through explicit relations in the query. The aliases, super-classes, sup-classes of entity in ontology are exploited to activate the spreading.

In the next section, we discuss background and related works. Section 3 describes the proposed system architecture and the model. Section 4 presents evaluation of the proposed model and discussion on experiment results in comparison to other models. Finally, section 5 gives some concluding remarks and suggests future works.

## 2. BACKGROUND AND RELATED WORKS

### 2.1. Ontology

There are many definitions of ontology being mentioned in the last years. In [10], the authors said that ontology was a formalization and explicit specification of a shared conceptualization. In that, formalization means that machine can understand, being explicit means that the type of concepts used and the constraints on their use are explicitly defined, and conceptualization is an abstract model that identifies phenomenon's relevant concepts. Two main application areas of ontology are information retrieval and knowledge management. There are various well-known ontologies, such as KIM ([21]) and YAGO ([35], [36]).

KIM (Knowledge and Information Management) is an infrastructure for automatic semantic annotation about named entity, indexing, and retrieval of documents. KIM ontology is an upper-level ontology containing about 300 concepts and 100 attributes and relations. The KIM keeps the semantic descriptions of entities in the KIM Knowledge Base (KB), which is repeatedly enriched with new entities and relations. KIM KB contains about 77,500 entities with more than 110,000 aliases. The entity descriptions are stored in the same RDF(S) repository. Each entity has information about its specific type, aliases, properties (attributes and relations).

YAGO (Yet Another Great Ontology) contains about 1.95 millions concepts, 93 different relation types, and 19 millions facts about specific relations between concepts. The facts are extracted from Wikipedia and WordNet using information extraction rules and heuristics. New facts are verified and added to the knowledge base by YAGO core checker. Therefore the correctness of the facts is about 95%. Note that, with logical extraction techniques, a flexible architecture, and open source, YAGO can be further extended in future or combined with some other ontologies.

### 2.2. Constrained Spreading Activation Model

In computing science, the SA model ([7]) was first used in the area of artificial intelligence, but recently it has been also widely used in information retrieval. The pure SA model consists of an associative network or a semantic network and simple processing techniques applied on the network to find concepts that are related to a user's query. The basic idea behind a SA algorithm is exploitation of relations between concepts in the network. The concepts may correspond to terms, documents, authors, and so forth. The relations are usually labelled and/or directed and/or weighted. The SA algorithm creates an initial set of concepts from the content of a user's query and assigns weights to them. After that, with the original concepts, it obtains a set of related concepts by navigating the relations in the network. The activated concepts are assigned weights and added into the original query. First, the concepts that are nearest to the original concepts are activated. After that, the activation reaches next level concepts through the networks using relation links. The process ends when one of termination conditions is reached.

The SA algorithm is a sequence of iterations. Each iteration consists of one or more pulses and a termination check. A pulse is created from three phases: (1) pre-adjustment; (2) spreading; (3) post-adjustment. In that, pre-adjustment and post-adjustment are optional which use some form of activation decay to control the concepts in the network. Spreading is activation waves from one concept to all other concepts connected to it. This is pure SA model which has some drawbacks. The most of them is meaningless concept activation by quickly spread over the network.

The drawbacks of the pure SA model can be partially surmounted by using some rules. Specially, the spreading of the activation could be combined with some constraints being inferences or heuristics. This new model is called constrained spreading activation (CSA). Some popular constraints are distance constraint, fan-out constraint, path constraint, activation constraint.

## 2.3. Systems using Spreading Activation

Due to the effectiveness of syntactic search appear some limitations. Semantic search is a major topic of current IR research. There are many technologies to improve effectiveness of search engine. In the scope of this paper, we only survey papers using SA algorithm. Because, pure-SA would return most results independent with queries ([3]). With the desire to improve the performance of document retrieval systems, many systems use SA algorithm constrained by some methods. However, there are no systems that use relations in given query to constrain SA algorithm. The followings are recent typical SA systems:

In [1], the system used a two-level SA network to activate strongly positive and strongly negative matches based on keyword search results. The system also used synonyms of original concepts of a query to activate, and the support vector machine method to train and classify the above data.

In [29], the authors proposed a hybrid spread activation algorithm that combined SA algorithm together with ontology based information retrieval. The algorithm enabled the user to represent his queries in keywords, and found concepts in the ontology having the keywords occur in their descriptions. The found concepts were counted as initial concepts and weights were assigned to links based on certain properties of the ontology, to measure the strength of the links. After that, the SA algorithm was used to find related concepts in ontology. The activated concepts did not contain any of the specified keywords.

In [18] and [19], the authors did not require users to specify the concepts of their queries. The system mapped the original query to a keyword set and searched for documents relating to the keyword set. After that, the documents were pre-annotated with information of the ontology and the initial concepts were extracted from the retrieved documents. An SA algorithm was used to find concepts semantically relating to the concepts in the ontology. Finally, the activated concepts were used to re-rank the documents to present for user.

In [30], the system found answers of given query and added into the query. After that, the system used an SA algorithm to find concepts that were related to the expanded query. Besides, [17] expanded query by using SA on all relations in WordNet and only selecting kernel words that were activated and represented the content of a query by some rules.

In [23], the system set up an associative network with nodes being web pages and links between the nodes being relations between the web pages. Initial nodes of SA algorithm are web pages that are strongly associated to given query. Next, other nodes (web pages) of their network are activated.

The above papers using SA method are not used explicit relations in query to constrain the spreading. Hence, the concepts added to the query can not really relevant to the query. In [37], the authors used the relations in the query to expand it. However, the paper only exploited spatial relations, such as *near*, *inside* and *north of*. Meanwhile, the our proposed method is not limited in the type of relation in query expansion.

## 3. QUERY-ORIENTED-CONSTRAINED SPREADING ACTIVATION

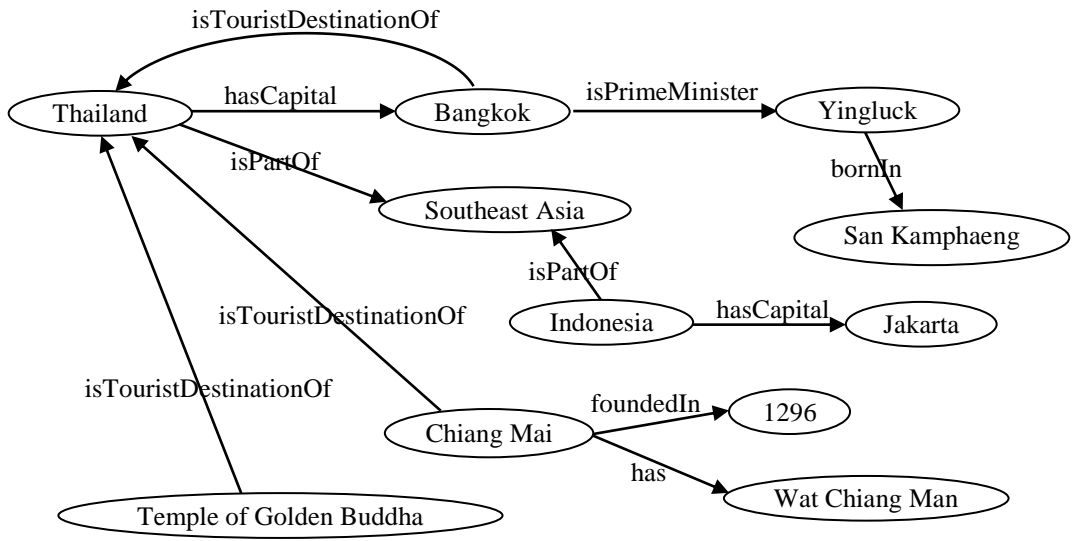

**Fig. 3.1.** Example about concepts related with the concept *Thailand* in a fact ontology

The weakness of the pure SA method can be partially overcome by using a limited number of rules about spreading. In constrained spreading activation, the spreading can be constrained by distance, fan-out, path and activation. Figure 3.1 illustrates a portion of a fact ontology containing concept *Thailand*. With a query searches documents about "*cities that are tourist destinations of Thailand*", based on the content of the query and events described in Figure 3.1, only two concept *Bangkok* and *Chiang Mai* should be activated and added to the query. Meanwhile, in the pure SA method, from initial concept *Thailand*, ten concepts *Bangkok*, *Yingluck*, *San Kamphaeng*, *Southeast Asia, Indonesia, Jakarta, Chiang Mai*, *1296*, *Wat Chiang Man*, and *Temple of Golden Buddha* will be activated and added to the query. There are eight inconsistent concepts are added to the query.

Besides, in the constrained spreading activation about distance is 1, i.e. only the concepts have direct relation with the original concept, four concepts *Bangkok*, *Southeast Asia*, *Chiang Mai*, and *Temple of Golden Budda* are activated and added to the query. However, *Southeast Asia* and *Temple of Golden Budda* are not consistent because *Southeast Asia* is not a city and *Temple of*

*Golden Budda* is a tourist destination, but is not a city of Thailand. Hence, we propose a Query-Oriented-Constrained Spreading Activation algorithm (QO-CSA) to overcome the weaknesses.

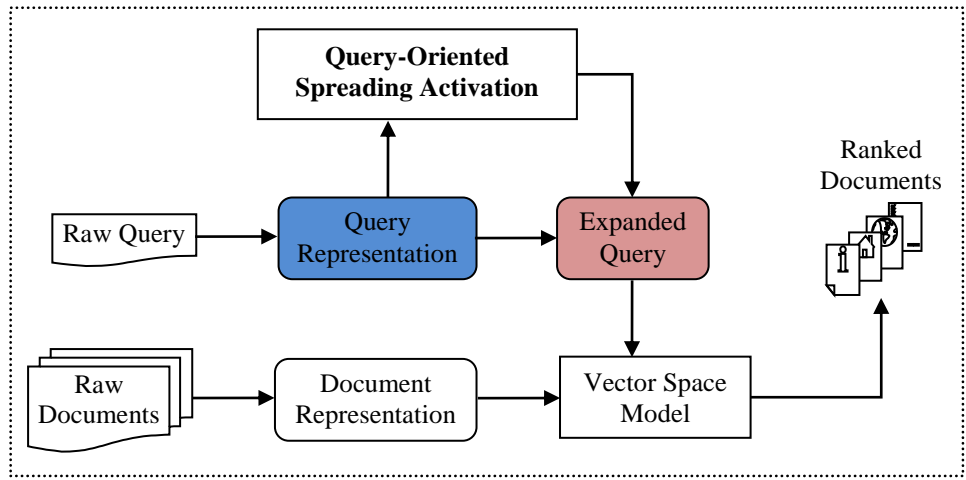

**Fig. 3.2.** System architecture for semantic search

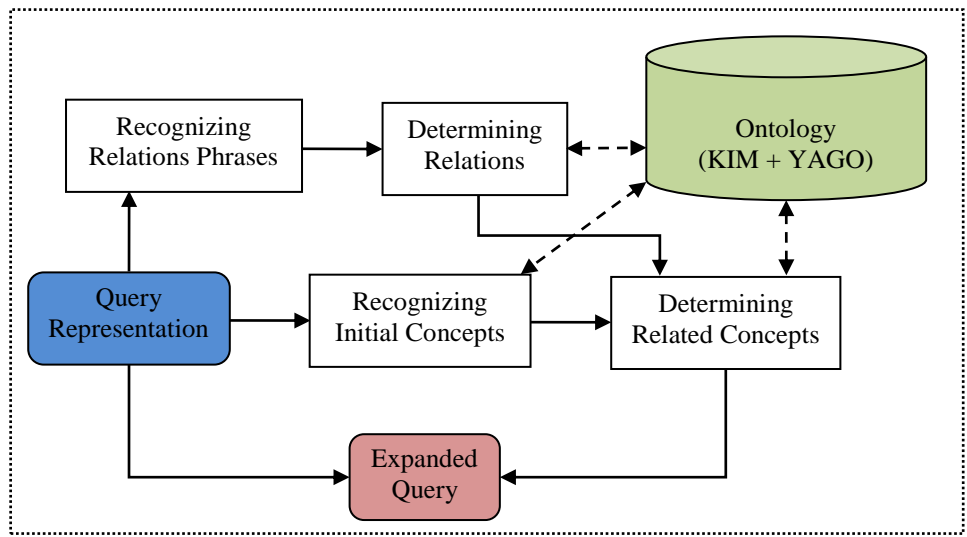

**Fig. 3.3.** The steps determining latently related entities for a query

Our proposed system architecture is shown in Figure 3.2. The Document Representation module extracts, stems and indexes keywords of raw documents. This module is implemented by Lucene ([13]). The Query Representation module represents a user's query by keywords, which are counted as initial concepts of QO-CSA module. After that, the QO-CSA module exploits the ontology to find really related concepts based on the initial concepts and the relations represented by the original query. This is presented specifically in section 3.2. Expanded query is inclusion of the original keywords and new keywords being the activated concepts. Semantic document search is performed via the Vector Space Model module presenting in section 3.3.

The query-oriented-constrained spreading activation is represented in Figure 3.3. The followings are the four main steps of our method to determine really latently related concepts for a query:
1. Recognizing Relation Phrases: Relation phrases are prepositions, verbs, and other phrases representing relations, such as *in*, *on*, *of*, *has*, *is*, *are*, *live in*, *located in*, *was actress in*, *is author of*, and *was born*. We have implemented a relation phrase recognition using the

ANNIE tool of GATE ([8]).

2. Determining Relations: Each relation phrase recognized in step 1 is mapped to the corresponding relation in fact ontology or NE ontology by a manually built dictionary. For example, "*was actress in*" is mapped to *actedIn*, "*is author of*" is mapped to *wrote*, and "*nationality is*" is mapped to *isCitizenOf*.

3. Recognizing Initial Concepts: we find concepts in the query by mapping the words expressed in the query to entity names or word forms in the exploited ontologies. These are original concepts in the query and initial concepts of the method.

4. Presenting each relation in the query in the relation form *I-R-C* (or *C-R-I*), where *R* is a relation found in step 2, and *I* and *C* are initial concepts (entity and class) found in step 3.

    Example, with the query "*Where is the actress, Marion Davies, buried?*", the relation phrases are "*where*" and "*buried*" are mapped to the relation *R* "*buriedIn*", the entity *I* "*Marion Davies*" has class "*Woman*" and the word "*Where*" is mapped to class *C* "*Location*". Hence the relation form is [*I*: #*Marion_Davies*]-(*R*: *buriedIn*)-[*C*: *Location*].

5. Determining related concepts: with each relation form *I-R-C*, find the latent entities $I_a$ having relation *R* with *I*, and having class being *C* or being sub-class of *C*.

    Example, in exploited ontology, there is a relation: [*I*: #*Marion_Davies*]-(*R*: *buriedIn*)-[$I_a$: #*Hollywood_Cemetery*] and #*Hollywood_Cemetery* is an entity having class being sub-class of *Location*. Hence, #*Hollywood_Cemetery* is a suitable entity for the relation form in step 4.

6. Before being added into the query, the latent concepts are represented by their main entity aliases. In above example, "*Hollywood Cemetery*" is added to the query.

The QO-CSA uses only activation from original nodes to all other nodes which connected to them and the connections are presented in the query. Comparing with pure-SA algorithm, the QO-CSA algorithm has two constraints following: (1) distance constraint: only nodes having links with original nodes in the exploited ontology are activated; (2) query constraint: the links is must presented in the query.

Our proposed model needs an ontology having: (1) a comprehensive class catalog with a large concept population for expressing clearly information of documents and queries; and (2) a comprehensive set of relations between concepts and facts for expanding queries with latently related concepts. Since no single ontology is rich enough for every domain and application, merging or combining multiple ontologies are reasonable solutions ([6])). So we have combined 2 ontologies KIM and YAGO to have a rich ontology for our model.

We use the vector space model (VSM) to match documents and queries, and rank retrieved documents. Each document or query is represented by a single vector over a keyword space. Vectors of the query q and the j[th] document are the following representations: $\vec{q} = (w_{1,q}, w_{2,q}, \ldots, w_{t,q})$, $\vec{d}_j = (w_{1,j}, w_{2,j}, \ldots, w_{t,j})$, with *t* is amount of keyword in corpus. The correlation of the document and the query is usually the cosine of the angle between the two vectors $\vec{d}_j$ and $\vec{q}$.

$$\text{sim}(d_j, q) = \frac{\vec{d}_j \bullet \vec{q}}{|\vec{d}_j| \times |\vec{q}|} = \frac{\sum_{i=1}^{t} w_{i,j} \times w_{i,q}}{\sqrt{\sum_{i=1}^{t} w_{i,j}^2} \times \sqrt{\sum_{i=1}^{t} w_{i,q}^2}}$$

Where $w_{i,j}$ and $w_{i,q}$ are weights of the keyword $k_i$ in document $d_j$ and query q. The weights were calculated as the traditional *tf.idf* scheme.

The main advantages of the vector space model include: (1) its non-binary term-weight improves retrieval performance; (2) its cosine ranking sorts documents according to their degrees of correlation to information request; and (3) it is simple and fast. For these reasons, the vector space model is a popular retrieval model nowadays ([1]).

We implement the above VSM by employing and modifying Lucene which is a general open source for storing, indexing and searching documents [13]. In fact, Lucene uses the traditional VSM with a tweak on the document magnitude in the cosine similarity formula for a query and a document. That however does not affect the ranking of the documents.

## 4. EXPERIMENTS

In this section, we make experiments about search performance of our system and comparing with the IR systems using SA technology. Evaluation of an IR model requires two components: (1) a test dataset including one document collection, one query collection and relevance information about which documents are relevant to each query; and (2) quality measures based on relevant and non-relevant documents returned for each query ([2], [27]). So, our system can compare with other systems by using one of following two methods:

1. Choose suitable a dataset and quality measures. After that, our system and the other systems will be evaluated on the dataset and the measures. Because the authors of the other systems do not often provide sources of the systems, we will reproduce them. However, some technical features in the systems are not described in their papers.

2. Our system will use with their dataset and measures to experiment. Because each systems use different dataset and measures, we must compare with each systems on same its dataset and measures. However, some systems use datasets built by the authors and the authors do not provide the datasets.

Hence, direct comparison is extremely difficult. Beside, in this paper, we only want present that our QO-CSA algorithm is better than CSA algorithm of the other systems. Our system will be indirectly compared with the other systems by: (1) choose a popular dataset and popular measures, which are often used by text IR community; and (2) reproduce system using CSA algorithm to compare with our system.

### 4.1. Dataset and Performance Measures

We choose the TREC Disk-5 document collection of the TREC datasets. This document collection is used widely in text IR community. Because among 56 papers about text IR of SIGIR-2007[1] and SIGIR-2008[2], there are 33 papers using TREC datasets and 15 papers using the TREC Disk-5. The TREC Disk-5 consists of more than 260,000 documents in nearly 1GB. Next, we choose queries of the QA-Track that have answer documents in this document collection. As presented in Table 4.1, among the queries of the QA-Track, there are only 858 queries of QA-99, QA-00, QA-01 that have answer documents in the TREC Disk-5. However, we only choose queries which can be expanded by our QO-CSA algorithm. There are 71 queries which contain at least one relation phrase that has a respective relation type and respective facts in the ontology of discourse.

---

[1] http://www.sigir2007.org
[2] http://www.sigir2008.org

Table 4.1. Query survey of the TREC QA-Track

| QA-Track | 1999 | 2000 | 2001 | Total |
|---|---|---|---|---|
| Number of queries | 200 | 693 | 500 | 1393 |
| Number of queries having answers in TREC Disk-5 | 155 | 441 | 262 | 858 |
| Number of queries expanded by our QO-CSA algorithm | 21 | 23 | 27 | 71 |

Using the chosen dataset, we evaluate the performance of the proposed model and compare it with others by the common precision measure (P), recall (R) measure and F-measure ([2], [27]). In addition, the experiment systems rank documents regarding their similarity degrees to the query. Hence, P-R curves represent the retrieval performance better and allow comparison of those of different systems. The closer the curve is to the right top corner, the better performance it represents. The average P-R curve and average F-R curve over all the queries are obtained by P-R curves and F-R curves of each query which are interpolated to the eleven standard recall levels that are 0%, 10%, …, 100% ([22], [2], [27]). Besides, MAP is also used to compare the systems because MAP is a single measure of retrieval quality across recall levels and considered as a standard measure in the TREC community ([37]).

Obtained values of the measures presented above might occur by chance due to: (1) the specific and restricted contents of queries and documents in a test dataset; (2) the subjective judgment of human assessors on relevance between the test queries and documents; and (3) the limited numbers of queries and documents used in an evaluation experiment. So, when comparing systems, a typical null hypothesis is that they are equivalent in terms of performance, even though their quality measures appear different. In order to reject this null hypothesis and confirm that one system truly performs better than another, a statistical significance test is required ([16]). We use Fisher's randomization (permutation) test for evaluating the significance of the observed difference between two systems, as recommendation of [32]. As shown [32], 100,000 permutations were acceptable for a randomization test and the threshold 0.05 of the two-sided significance level, or two-sided p-value, could detect significance.

**4.2. Testing results**

We conduct experiments to compare the results obtained by three different search models:

1. Query-Oriented-Constrained Spreading Activation Search (QC-OSA Search): This is the search engine that uses the proposed model represented in section 3.

2. Syntactic Search: This is the Lucene text search engine as a tweak of the traditional keyword-based VSM.

3. Constrained Spreading Activation Search (CSA Search): This is the search engine using the traditional constrained SA algorithm, which is similar to QC-OSA Search but it expands query by broadcasting all links to find related concepts. Similarly to QC-OSA Search, CSA Search is constrained about distance. CSA Search only activates once. Specifically, it activates nodes having links with initial nodes of the query.

**Table 4.2.** The average precisions and F-measures at the eleven standard recall levels on 71 queries of the TREC Disk-5

| Measure | Model | Recall (%) | | | | | | | | | | |
|---|---|---|---|---|---|---|---|---|---|---|---|---|
| | | 0 | 10 | 20 | 30 | 40 | 50 | 60 | 70 | 80 | 90 | 100 |
| Precision (%) | Syntactic Search | 56.8 | 56.7 | 52.6 | 49.3 | 45.3 | 44.5 | 36.2 | 30.7 | 27.9 | 25.9 | 25.1 |
| | CSA Search | 47.4 | 45.1 | 43.2 | 39.6 | 36.6 | 35.3 | 28.7 | 25.6 | 23.8 | 23.0 | 22.5 |
| | QO-CSA Search | 65.4 | 65.3 | 61.5 | 57.7 | 53.5 | 51.6 | 42.3 | 37.5 | 34.3 | 33.3 | 32.6 |
| F-measure (%) | Syntactic Search | 0.0 | 14.7 | 24.4 | 31.2 | 35.0 | 38.9 | 35.7 | 33.2 | 31.7 | 30.4 | 30.1 |
| | CSA Search | 0.0 | 11.9 | 19.7 | 24.6 | 27.7 | 30.2 | 28.0 | 26.8 | 25.9 | 25.9 | 26.3 |
| | QO-CSA Search | 0.0 | 15.3 | 26.1 | 33.7 | 38.5 | 42.7 | 39.9 | 38.2 | 36.3 | 37.0 | 37.5 |

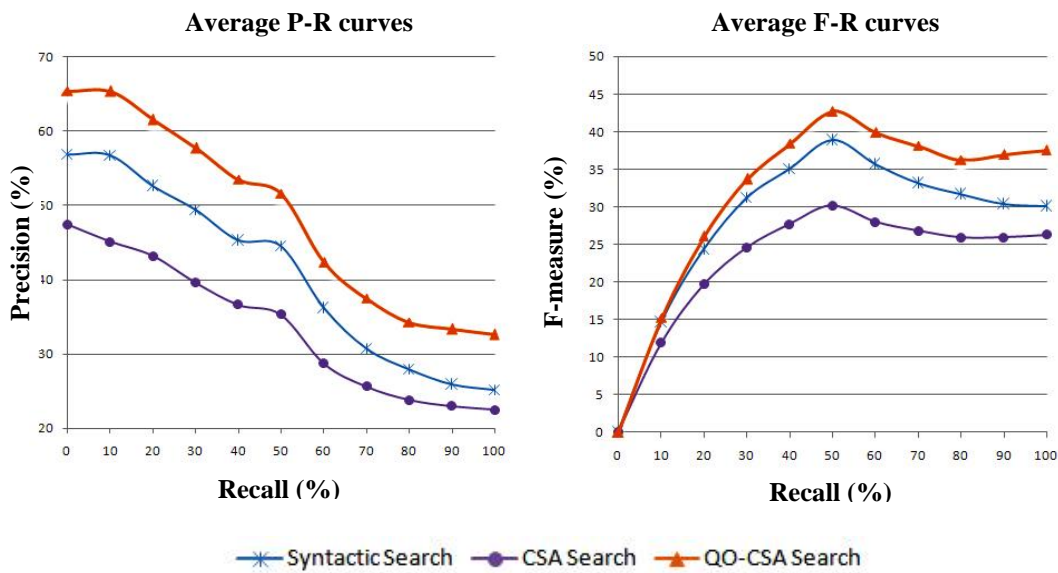

**Fig. 4.1.** Average P-R and F-R curves of Syntactic Search, CSA, QO-CSA models on 71 queries of TREC

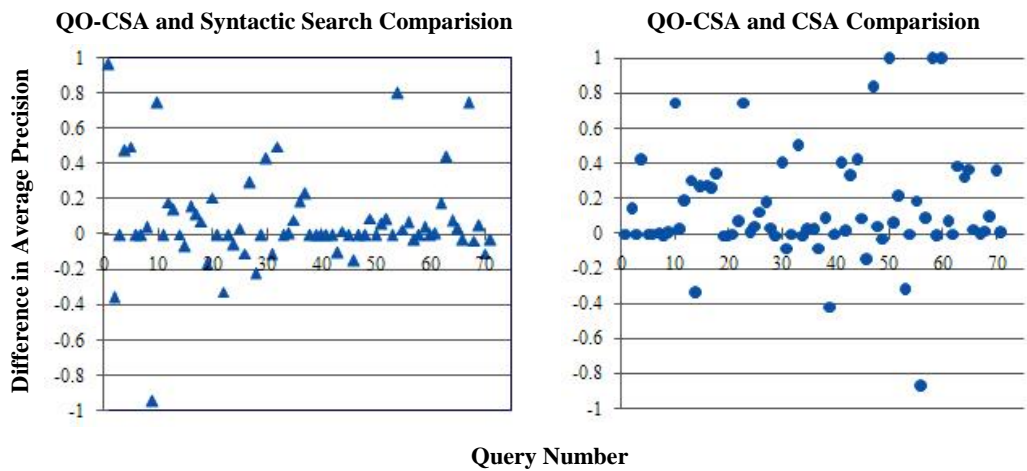

**Fig. 4.2.** The per query differences in average precision of QO-CSA Search with the Syntactic Search and CSA Search models

Table 4.2 and Figure 4.1 plots present the average precisions and F measures of Syntactic Search, CSA Search and QO-CSA Search models at each of the standard recall levels. It shows that QO-CSA Search performs better than the other two models, in terms of the precision and F measures. Figure 4.2 shows the per query differences in average precision of QO-CSA Search with the Syntactic Search and CSA Search models. The MAP values in Table 4.3 and the two-sided p-values in Table 4.4 show that queries expanded by really suitable information enhance text retrieval performance. In terms of the MAP measure, the QO-CSA Search performs about 18.9% and 43.8% better than the syntactic search and the CSA search, respectively.

Table 4.3. The mean average precisions on the 71 queries of TREC

| Model | QO-CSA Search | Syntactic Search | CSA Search |
|---|---|---|---|
| **MAP** | 0.4703 | 0.3957 | 0.3271 |
| **Improvement** | | 18.9% | 43.8% |

Table 4.4. Randomization tests of QO-CSA Search with the Syntactic Search and CSA Search models

| Model *A* | Model *B* | \|MAP(*A*) – MAP(*B*)\| | $N^-$ | $N^+$ | Two-Sided P-Value |
|---|---|---|---|---|---|
| **QO-CSA Search** | **Syntactic** | 0.0746 | 2129 | 2402 | 0.04531 |
| | **CSA Search** | 0.1432 | 1710 | 1664 | 0.03374 |

## 5. CONCLUSION AND FUTURE WORKS

We have presented the VSM that uses query-oriented-constrained spreading activation to exploit really related concepts for semantic text search. That is a whole IR process, from a natural language query to a set of ranked answer documents. The conducted experiments on a TREC dataset have showed that our appropriate ontology exploitation improves the search quality in terms of the precision, recall, F, and MAP measures. Although this work uses VSM for proving the advantage of discovering latent concepts in text retrieval, it could be adapted for other information retrieval models as well.

For future work, we are considering combination of YAGO, KIM, WordNet and other ontologies to have a great ontology to increase the relation coverage and researching methods to better recognize relations in a query. Besides, the our system need exploits and annotates ontological features of named entities and WordNet words in documents and queries for semantic text search.